\documentclass[10pt,twocolumn,letterpaper]{article}

\usepackage{cvpr}              

\usepackage{graphicx}
\usepackage{amsmath}
\usepackage{amssymb}
\usepackage{booktabs}
\usepackage{enumitem}
\usepackage{caption}

\usepackage[accsupp]{axessibility}
\usepackage{times}
\usepackage{epsfig}

\usepackage[pagebackref,breaklinks,colorlinks]{hyperref}

\usepackage[capitalize]{cleveref}
\crefname{section}{Sec.}{Secs.}
\Crefname{section}{Section}{Sections}
\Crefname{table}{Table}{Tables}
\crefname{table}{Tab.}{Tabs.}

\def\cvprPaperID{171} 
\def\confName{CVPR}
\def\confYear{2022}

\begin{document}

\title{Structure-Aware Motion Transfer with Deformable Anchor Model}

\author{Jiale Tao$^1$\footnotemark[1] \quad Biao Wang$^2$ \quad Borun Xu$^1$\footnotemark[1] \quad Tiezheng Ge$^2$ \quad Yuning Jiang$^2$ \quad Wen Li$^1$ \quad Lixin Duan$^1$
\vspace{3pt}\\
\normalsize$^1$School of Computer Science and Engineering \& Shenzhen Institute for Advanced Study,\\
\normalsize University of Electronic Science and Technology of China\\
\normalsize$^2$Alibaba Group\\
\tt\small \{jialetao.std, liwenbnu, lxduan\}@gmail.com, xbr\_2017@std.uestc.edu.cn\\
\tt\small \{eric.wb, tiezheng.gtz, mengzhu.jyn\}@alibaba-inc.com
}

\twocolumn[{%
\renewcommand\twocolumn[1][]{#1}%
\maketitle
\begin{center}
    \centering
    \includegraphics[width=1\linewidth,height=0.65\columnwidth]{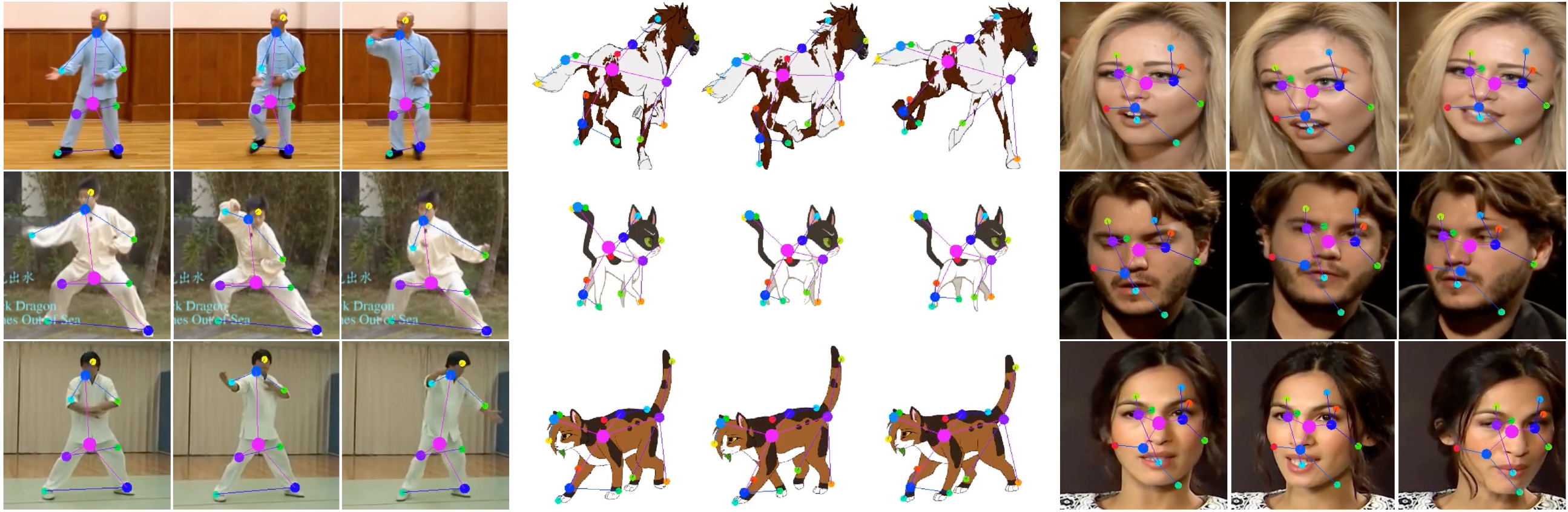}
    \captionof{figure}{Examples of structures learned by our proposed approach for the motion transfer task. We present results on three different datasets: TaiChiHD, MGIF and VoxCeleb1. It is worth noting that no prior structural information (\eg, skeleton) is used in our approach.}
    \label{fig:introduction}
\end{center}%
}]

\renewcommand{\thefootnote}{\fnsymbol{footnote}}
\footnotetext[1]{Work done during an internship at Alibaba Group}
\footnotetext[2]{Codes will be available at https://github.com/JialeTao/DAM.git}

\begin{abstract}

Given a source image and a driving video depicting the same object type, the motion transfer task aims to generate a video by learning the motion from the driving video while preserving the appearance from the source image. In this paper, we propose a novel structure-aware motion modeling approach, the deformable anchor model (DAM), which can automatically discover the motion structure of arbitrary objects without leveraging their prior structure information. Specifically, inspired by the known deformable part model (DPM), our DAM introduces two types of anchors or keypoints: i) a number of motion anchors that capture both appearance and motion information from the source image and driving video; ii) a latent root anchor, which is linked to the motion anchors to facilitate better learning of the representations of the object structure information. Moreover, DAM can be further extended to a hierarchical version through the introduction of additional latent anchors to model more complicated structures. By regularizing motion anchors with latent anchor(s), DAM enforces the correspondences between them to ensure the structural information is well captured and preserved. Moreover, DAM can be learned effectively in an unsupervised manner. We validate our proposed DAM for motion transfer on different benchmark datasets. Extensive experiments clearly demonstrate that DAM achieves superior performance relative to existing state-of-the-art methods.
\end{abstract}

\section{Introduction}

Recently, motion transfer has gained increasing attention from computer vision researchers, due to its numerous potential applications in the fields of video re-enactment~\cite{chan2019everybody}, fashion design~\cite{dong2019fw}, face swapping~\cite{siarohin2020first}, and so on. Given a source image and a driving video of the same object type, the goal of motion transfer is to generate a video that depicts the motion pattern contained in the driving video while preserving the appearance from the source image.

\begin{figure}[t]
\begin{center}
   \includegraphics[width=0.88\linewidth,height=0.6\columnwidth]{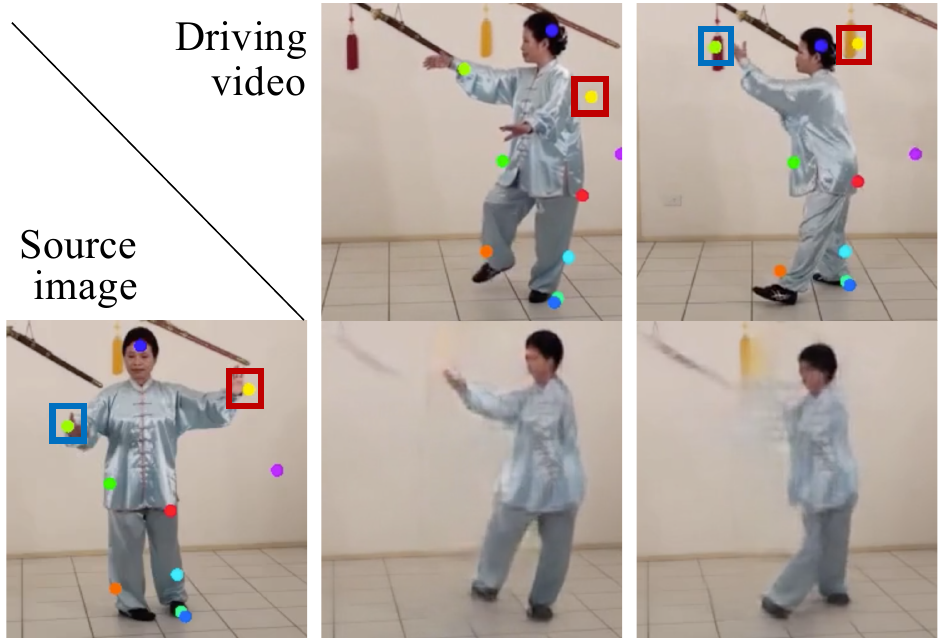}
   \caption{Failure cases from the FOMM method~\cite{siarohin2020first}. Inaccurate correspondences between motion points cause parts of the human body to be missed in the generated videos.}
   \label{fig:background prediction}
  \vspace{-0.6cm}
\end{center}
\end{figure}

Finding the correspondence between a source image and a driving video is the key to successful motion transfer. Existing motion transfer methods address this issue in two ways. On one hand, model-based methods~\cite{ma2017pose,gu2020flnet} utilize a pre-trained third-party model to extract the structural information of an object (\eg, human bodies, human faces, etc.). However, specific predefined structure priors are required for different objects. On the other hand, model-free methods~\cite{siarohin2019animating,siarohin2020first,wiles2018x2face} treat motion keypoints as unknown variables, then design models to predict them by optimizing the image reconstruction loss. While these approaches do not require a predefined object structure, they often suffer from false correspondences, leading to considerable artifacts emerging in the generated videos (see Fig.~\ref{fig:background prediction} for examples).

To address these issues, in this paper, we propose a novel structure-aware motion transfer approach referred to as the deformable anchor model (DAM). In DAM, we take advantage of both the model-free and model-based methods. On one hand, similar to the model-free methods, we represent motion keypoints (a.k.a., ``anchors") as unknown variables, which enables our model to perform motion transfer on an arbitrary object without knowing its prior structural information. On the other hand, to prevent the false correspondences, we also encode the structural information to constrain those motion anchors. Unlike model-based methods, our approach does not employ any pre-trained third-party model. Instead, as inspired by the well-known deformable part model (DPM)~\cite{DPM}, DAM introduces a latent root anchor to regularize the motion anchors and model the object structure, enabling the correspondence between the source image and driving video to be enforced and thus further improving the performance. Furthermore, by introducing additional latent anchors, DAM can be easily extended to a hierarchical version that can more effectively model complicated object structures. 
Note that all latent anchors in our DAM are unknown variables, and that DAM can be learned in an end-to-end manner, similarly to previous model-free methods. 

We conduct experiments on four benchmark datasets (\ie, TaiChiHD, FashionVideo, VoxCeleb1 and MGIF) for performance evaluation. The experimental results show that our method not only achieves the best quantitative performance, but also exhibits a strong capacity to capture the motion structure of different objects, such as human bodies, faces, animals, and so on.


\section{Related Work}

\noindent{\bf Video-to-video synthesis:} Motion transfer has been studied in the area of video-to-video synthesis to some extent. Video2video~\cite{wang2018video} proposed to synthesize photo-realistic videos via input video semantic maps. Chan \etal\cite{chan2019everybody,yang2020transmomo} further extended the generation scheme to synthesize human dance videos conditioned on input video pose sequences and a source identity. These methods are good at utilizing the input source appearance information and can generate realistic videos. However, they are also identity-specific methods, meaning that they require a large amount of source images with diverse views and ranges of motion and moreover take a long time to train. 

\noindent{\bf Motion transfer:} Early methods\cite{ma2017pose,ma2018disentangled,balakrishnan2018synthesizing,siarohin2018deformable,wei2020c2f} mainly focus on pose-guided human image generation. These works use off-the-shelf pose estimators or keypoint detectors to pre-extract pose information, which is then adopted for conditioning the image generation process. A series of works~\cite{zhu2019progressive,chen2019unpaired,li2019dense,ren2020deep,neverova2018dense,liu2019liquid, ren2021flow,zhang2021pise,Sarkar2020,Yoon_2021_CVPR,pumarola2018ganimation} have adopted this approach. In addition, many works have proposed facial animation methods~\cite{wei2020learning,gu2020flnet,tripathy2021facegan,burkov2020neural,chen2020puppeteergan,wang2021one,yao2021one,kim2018deep} which can be seen as a kind of facial motion transfer. Similarly, these methods also employ an off-the-shelf facial landmark detector for expression modeling. 

Despite their ability to transfer the pose of a human body or the expression on a human face, these methods heavily rely on third-party models and are object-specific. Inspired by Jakab et al.~\cite{jakab2018unsupervised}, which proved that object landmarks can be learned in an unsupervised way via image reconstruction, Monkey-Net~\cite{siarohin2019animating} was the first to propose a model-free motion transfer method for arbitrary objects, which was achieved by building backward motion flow from aligned keypoints to warp the source image feature to driving pose. This warping-based method can achieve superior motion modeling and transferring performance, but this performance begins to suffer when the motions in question are large and complex. FOMM~\cite{siarohin2020first} enhances the motion model by introducing local affine transformations to motion keypoints. Since no structural information is provided, however, this approach often suffers from unstable correspondence between the source and driving image. RegionMM~\cite{PCAMotion} further extends the FOMM by defining regions that can be used to model parts of an object, although it does not consider the dependent structure between the different regions.

\noindent{\bf Other related work:} Most of the above motion transfer methods rely on keypoint detection for encoding pose information. Generally speaking, model-based methods tend to adopt supervised keypoint detection or pose estimation methods~\cite{cao2017realtime,newell2016stacked,zhang2015learning,yu2016deep}, while model-free methods tend to be unsupervised keypoint detection methods~\cite{zhang2018unsupervised,jakab2018unsupervised}. For supervised cases, keypoints are learned on additional and richly annotated datasets. For unsupervised cases, keypoints are usually learned via an auxiliary image reconstruction task. Specifically, detected keypoints are considered to represent the structural information of an image object; the image should be reconstructed via combining the structural and appearance information.

Our work is partially inspired by DPM~\cite{DPM}, which is a traditional human object detection approach. It breaks down the task into individual part detection task across human body and defines the score of positive detection of a root location by considering the spatial distance prior between root location and part locations. Intuitively, if the current relative distance from a part (\eg the left leg of a human) to the root (\eg the head of a human) is much larger than the prior relative distance, then these root-part pair locations tend to be assigned a lower positive detection score in DPM. In a similar spirit, we consider the motion prior of a root anchor which is formulated in a similar way to the spatial distance constraint in DPM.

\section{Structure-Aware Motion Transfer}

In this section, we present our structure-aware motion transfer approach which we name the deformable anchor model (DAM). We develop our approach based on the recent first-order motion model (FOMM~\cite{siarohin2020first}), with the addition of two novel deformable anchor models to encode the motion structure information. Below, we first present a brief review of the FOMM method in Section~\ref{3.1}, and then introduce the basic deformable anchor model in Section~\ref{3.2}. A more effective hierarchical deformable anchor model is presented in Section~\ref{3.3}, followed by a summary of the entire model in Section~\ref{3.4}.

\subsection{Motion Flow Modeling}\label{3.1}

Given a source image and a driving video, FOMM~\cite{siarohin2020first} generates the motion transfer video by warping the source image to mimic the driving video in a frame-by-frame manner. For this purpose, they firstly estimate the dense motion flow between these two images. They then warp the source image in the feature space, and synthesize the driving frame with an image generator based on the warped source image feature. The entire process is illustrated in the bottom part of Figure~\ref{fig:pipeline}.

Formally, given a source image $S$ and a driving frame $D$, the motion between two images is modeled by the motion flow $\mathcal{T}_{S \leftarrow D} (z)$, where $z$ denotes the coordinates of any pixel in the image. Estimating the dense motion flow is nontrivial. To ease this process, FOMM employs a set of motion anchors; these anchors are intended to represent identical keypoints of the object in the source image and driving frame (for example, the corresponding physical parts of the human body). With the aid of aligned motion anchors the dense motion flow can be derived through affine transformations. 

In more detail, let $z^s_k$ and $z^d_k$ denote the $k$-th pair of corresponding anchors in $S$ and $D$ respectively; here, $k=1,\ldots, K$, where $K$ is the number of motion anchors. This yields the following:
\begin{equation}
\begin{split}
z^s_k = \mathcal{T}_{S \leftarrow D} (z^d_k) \label{motion anchor point flow}
\end{split}
\end{equation}

Given a motion anchor, the motion flow for pixels at the local region around the anchor can be approximately modeled with an affine transformation. For convenience, let $\mathcal{T}_k$ denote the dense motion flow derived by the $k$-th motion anchor. The affine transformation can thus be described as follows:
\begin{equation}
\mathcal{T}_k (z) = \mathcal{T}_k (z^d_k) + \theta_k (z-z^d_k) \label{motion anchor flow}
\end{equation}
where $\theta_k$ is the parameter of local affine transformation for the $k$-th anchor. 

Intuitively, the dense motion flow of a pixel $z$ can be derived from any nearby motion anchor. Thus a weight parameter $M_k (z)$ is introduced to automatically combine the $\mathcal{T}_k (z)$ from different anchors. The dense motion flow of any pixel $z$ can thus be represented as follows:
\begin{equation}
\mathcal{T}_{S \leftarrow D} (z) = \sum_{k=1}^{K}M_k (z)\cdot\mathcal{T}_k (z), \label{global flow}
\end{equation}
where $\sum_{k=0}^{K}M_k (z) =1, \forall z$, in which $M_0 (z)$ is an additional mask for modeling background similar to the approach~\cite{PCAMotion}.

With $z^d_k$,$z^s_k$,$\theta_k$, and $M_k$, it is possible to obtain a dense motion flow between the source image $S$ and driving frame $D$, after which the source image can be warped to mimic the driving frame with an image generator. By enforcing an image reconstruction loss on the image generator, a motion estimator can be trained to automatically predict these unknown variables (\ie, $z^d_k$,$z^s_k$,$\theta_k$, and $M_k$) (see Fig.~\ref{fig:pipeline}). 

As FOMM has shown, the motion anchors tend to have coarse physical meanings (\eg, a motion anchor may always locate at the head region of a human). However, false correspondences may occur if large motion or background variance are present, which will lead to considerable artifacts in the generated videos,
as shown in Fig.\ref{fig:background prediction}. We will discuss how to address these issues by encoding latent object structure information in the following subsections.  

\begin{figure}[t]
\begin{center}
  \includegraphics[width=0.9\linewidth]{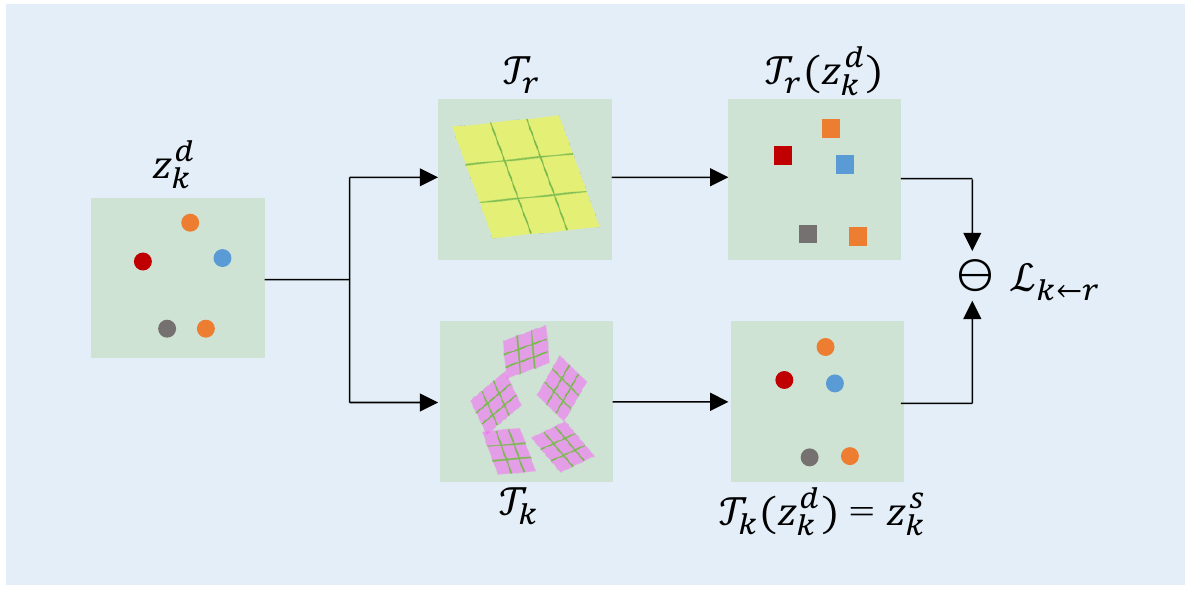}
  \caption{Illustration of Eqn.~(\ref{root2kp distance}). The colored squares denote prior flow derived from the root anchor, as described in Eqn.~(\ref{root flow at motion anchor point}), while the colored dots denote motion anchors. Euclidean distance is minimized between the pairs.}
  \label{fig:root prior}
  \vspace{-0.5cm}
\end{center}
\end{figure}

\subsection{Deformable Anchor Model (DAM)} \label{3.2}

\begin{figure*}[ht]
\begin{center}
   \includegraphics[width=1.0\linewidth]{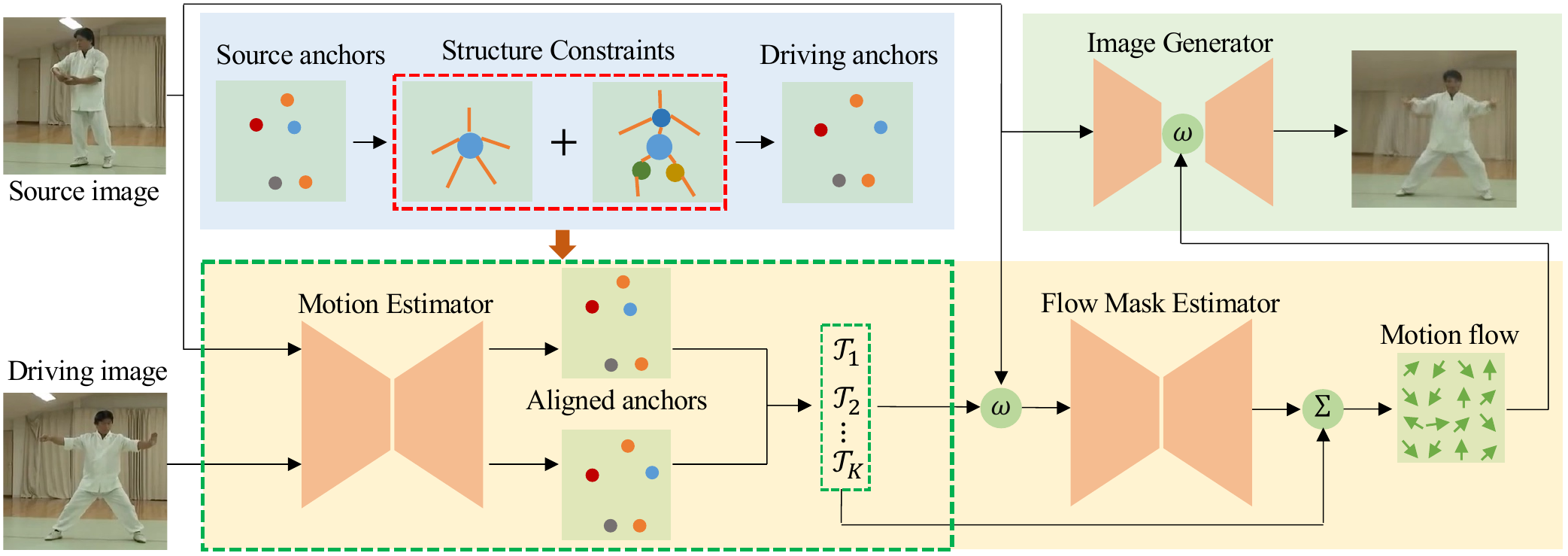}
    \caption{Overview of the proposed method. Anchors of the source image and driving image are respectively predicted through the motion estimator (we draw five motion anchors for clarity). The generated anchors are then are fed into a flow mask estimator together with the source image. The motion anchors and the flow masks are subsequently combined to obtain the dense warping flow for image generation. Note that motion anchors are constrained by the root anchor (\eg. the largest dot) and intermediate root anchors (\eg. medium-sized dots).}
    \label{fig:pipeline}
    \vspace{-0.5cm}
\end{center}
\end{figure*}

As discussed above, artifacts can be observed in the generated videos by FOMM. This is largely due to the motion anchors in FOMM are not properly regularized. Although different anchors are summed by the $M_k(z)$'s through Eqn~(\ref{global flow}), we observe that $M_k (z)$ tends to focus on only a local region around the anchor $z_k$ due to the assumption of affine transformation. As a result,  the $z_k^s$ and $z_k^d$ predicted by the motion estimator may not be accurately corresponded, leading to errors in the dense motion flow and artifacts in the generated video. 

To address this issue, we propose a new deformable anchor model (DAM) to discover the motion structure information of the object, then employ this information to regularize the motion anchors. In more detail, our model is inspired by DPM~\cite{DPM}. We introduce an additional \emph{latent root anchor} to establish communications among motion anchors. In a similar spirit to DPM, by connecting motion anchors with the root anchor, we expect the model to become aware of the motion structure of an object, even if its appearance varies in source images and driving videos.

Intuitively, given a source image and a driving frame, the root anchor represents the global motion between the two objects, which means that the flow of motion anchors should be related to that of the root anchor. Let $z^d_{r}$ denote the latent root anchor of the driving frame. We then model the relation between the motion and root anchors with an affine transformation, as follows:
\begin{equation}
    \mathcal{T}_r\left (z^d_{k}\right) = \mathcal{T}_r\left (z^d_{r}\right)+\theta_r \left (z^d_{k}-z^d_{r}\right)
    \label{root flow at motion anchor point}
\end{equation}
where $\mathcal{T}_r\left (z^d_{k}\right)$ is the derived flow based on the latent root anchor using the affine transformation model. We then regularize the motion flow of $z^d_k$ to be similar to the derived flow using the following loss:
\begin{equation}
   \mathcal{L}_{k\leftarrow r} = \left\|\mathcal{T}_k (z^d_k)-\mathcal{T}_r\left (z^d_{k}\right)\right\|_2 \label{root2kp distance}
\end{equation}

A further explanation of Eqn.~(\ref{root flow at motion anchor point}) and (\ref{root2kp distance}) is provided in Fig.~\ref{fig:root prior}. In a departure from the original FOMM, where the motion anchors are almost independent, we encode a latent object structure to regularize the motion anchors. Eqn.~(\ref{root flow at motion anchor point}) implies that we assume an affine transformation relation between the flow of the root anchor and motion anchors. While this may be stricter than required, divergence from ideal cases is permitted, and we use the derived flow as a prior to regularize the motion anchors with Eqn.~(\ref{root2kp distance}). 

On the other hand, through the use of Eqn.~(\ref{root flow at motion anchor point}) and (\ref{root2kp distance}), the motion anchors also guide us to learn a meaningful root anchor. As shown in Fig.~\ref{fig:ablation}, the root anchor is always located at the object centroid to capture the global movement of the object from one image to another. 

It should further be noted that, at the training stage, the latent root anchor $z^d_r$ and the affine transformation parameters $\theta_r$ can be obtained 
by the motion estimator in a similar way as the motion anchors. At the testing stage, the root anchor is discarded, and we only need to use the predicted motion anchors to generate dense motion flow in the same way as FOMM. The overall architecture of our method is illustrated in Fig.~\ref{fig:pipeline}.

\subsection{Hierarchical DAM} \label{3.3}
As discussed above, using an affine transformation to model the structure prior might be too restrictive, especially for objects with complicated motion. Taking the human body as an example, a movable part (\eg, left leg) might contain multiple joints, meaning that a single affine transformation can scarcely be expected to describe such a complex structure prior. 

This motivates us to construct a hierarchical deformable anchor model to facilitate the modeling of more complicated object structures. In more detail, we additionally introduce a set of latent intermediate anchors into the basic deformable anchor model. Rather than directly regularizing the motion anchors with the latent root anchor, we instead use latent intermediate anchors to regularize motion anchors and the latent root anchor to regularize latent intermediate anchors. Similarly, the affine transformation prior is applied between different types of anchors. Let $z^d_{i}$ denote an intermediate anchor; accordingly, we have:
\begin{eqnarray}
    \mathcal{T}_r\left (z^d_{i}\right) &=& \mathcal{T}_r\left (z^d_{r}\right)+\theta_r \left (z^d_{i}-z^d_{r}\right)\\
    \mathcal{T}_i\left (z^d_{k}\right) &=& \mathcal{T}_i\left (z^d_{i}\right)+\theta_i \left( z^d_{k}-z^d_{i}\right)
    \label{hdam_affine}
\end{eqnarray}
where $\theta_r$ and $\theta_i$ are affine transformation parameters of the root anchor $z^d_{r}$ and intermediate anchor $z^d_{i}$,  while $\mathcal{T}_r$ and $\mathcal{T}_i$ are the respective derived flows.

Accordingly, the loss for regularizing the motion flow of motion anchors can be written as follows: 
\begin{eqnarray}
\mathcal{L}_{k\leftarrow i} &=& \left\|\mathcal{T}_k (z^d_k)-\mathcal{T}_{i}\left (z^d_{k}\right)\right\|_2  \label{subroot2kp distance}\\
\mathcal{L}_{i\leftarrow r} &=& \left\|\mathcal{T}_{i} (z^d_{i})-\mathcal{T}_{r}\left (z^d_{i}\right)\right\|_2 \label{root2subroot distance}
\end{eqnarray}
Note that although the loss $\mathcal{L}_{k\leftarrow i}$ is defined for every pair of  $z^d_{i}$ and  $z^d_{k}$, we expect it takes effect on several  $z^d_{k}$'s nearby $z^d_{i}$ only. Therefore, in implementation, we assign attention weights to all $z^d_k$'s for each $z^d_{i}$, and allow the model to adjust these weights automatically.

With latent intermediate anchors, we can model a three-level hierarchical structure for object motion. To this end the procedure illustrated in Fig.~\ref{fig:root prior} can be further extended, where image pixels are involved, above which are the motion anchors, intermediate anchors and the root anchor respectively. By applying the affine transformation prior between the adjacent levels, we are able to model more complex object structures.

\subsection{Training DAM and HDAM} \label{3.4}
In both the basic deformable anchor model and hierarchical deformable anchor model, the newly introduced latent root anchors and the latent intermediate anchors can be predicted by the motion estimator network, which can be trained similarly to FOMM, \ie in an end-to-end fashion by optimizing the image reconstruction loss. 

More specifically, following FOMM~\cite{siarohin2020first}, we utilize the perceptual loss as our main driving loss, which is usually defined with a pre-trained VGG-19 networks~\cite{simonyan2014very}. Given a driving image $D$, the perceptual loss can be expressed as follows:
\begin{equation}
    \mathcal{L}_{per}=\frac{1}{C \cdot H \cdot W} \sum_{l}\left\|\phi_{l} (D)-\phi_{l}(\tilde {D})\right\|
\end{equation}
where $\tilde {D}$ is the generated driving image, $\phi_{l}$ denotes the feature extractor using the $l$-th layer of the VGG-19 network, and $C,H,W$ denote the number of channels, feature map height and width respectively.

Additionally, similar to recent works~\cite{siarohin2020first,zhang2018unsupervised}, an equivariance loss is adopted to ensure the geometric consistency of the learned anchors. For a known geometric transformation $\mathbf{T}$ and a given image $I$, the loss is defined as follows:
\begin{equation}
    \mathcal{L}_{equi}=\sum_{k}\left\|z^I_k-\mathbf{T}^{-1} (z^{\mathbf{T} (I)}_k)\right\|
    \label{equi loss}
\end{equation}

\textbf{Training DAM:} For the basic deformable anchor model, we write the loss for regularizing motion anchors as follows:
\begin{equation}
    \mathcal{L}_{dam} =  \sum_{k=1}^K\mathcal{L}_{k\leftarrow r}
    \label{fist order loss}
\end{equation}
where $\mathcal{L}_{k\leftarrow r}$ is defined in Eqn.~(\ref{root2kp distance}). 

The total training loss of our DAM model can be defined as:
\begin{equation}
    \mathcal{L} =  \mathcal{L}_{per} + \mathcal{L}_{equi} + \mathcal{L}_{dam} 
    \label{dam_loss}
\end{equation}
where we apply equal weights for all losses. 

\textbf{Training HDAM:}
For the hierarchical deformable anchor model, assuming a total of $I$ intermediate anchors are used, the loss can be written as follows:
\begin{equation}
    \mathcal{L}_{hdam} = \sum_{i}^{I}\left (\sum_{k=1}^K\omega_{ik}\mathcal{L}_{k\leftarrow i}+ \mathcal{L}_{i\leftarrow r}\right)
    \label{second order loss}
\end{equation}
where $\mathcal{L}_{k\leftarrow i}$ and $\mathcal{L}_{i\leftarrow r}$ are respectively defined in Eqn.~(\ref{subroot2kp distance}) and Eqn.~(\ref{root2subroot distance}); moreover, $\omega_{ik}$ denotes the attention weight between motion anchor $k$ and intermediate anchor $i$, which is computed through a fully connected layer. More detailed information about the attention process is presented in the supplementary material. 

The total training loss of our HDAM model can be defined as:
\begin{equation}
    \mathcal{L} =  \mathcal{L}_{per} + \mathcal{L}_{equi} + \mathcal{L}_{hdam} 
    \label{hdam_loss}
\end{equation}
where we also apply equal weights for all losses. In practice, when training HDAM, we use a pretrained DAM model as the initial model, then optimizing the loss in Eqn.~(\ref{hdam_loss}). 


\section{Experiments}

In this section, we evaluate our method on the benchmark datasets, and further provide insightful analysis by means of an ablation study and qualitative results. 

\subsection{Experimental Setup}
\noindent\textbf{Datasets:} We follow FOMM~\cite{siarohin2020first} and RegionMM~\cite{PCAMotion} in evaluating our method on four benchmark datasets containing different types of object:
\begin{itemize}[itemsep=0pt]
    \item TaiChiHD~\cite{siarohin2020first} contains 2867 training videos and 253 test videos. This dataset contains Tai-chi performers with different identities and various backgrounds, and is thought to be the most challenging dataset in this area due to its large motion. Two resolution variants of this dataset are evaluated: 1) all raw videos are cropped and resized to the basic $256\times 256$ resolution, as with FOMM; 2) the $512\times 512$ resolution, is a subset that removes any raw videos that fail to satisfy the resolution request for cropping, which contains 962 training videos and 112 testing videos.
    \item FashionVideo~\cite{zablotskaia2019dwnet} contains 500 training videos and 100 test videos. Videos in this dataset depict a single posing model with diverse clothing and textures. All videos are resized to a $256\times 256$ resolution.
    \item MGIF collected in~\cite{siarohin2019animating}, is a cartoon animal dataset containing 900 training videos and 100 test videos. All videos are resized to a $256\times 256$ resolution. 
    \item VoxCeleb1~\cite{nagrani2017voxceleb} is a talking head dataset, containing 19522 training 525 test videos. All videos are resized to a $256\times 256$ resolution. 
\end{itemize}

\noindent\textbf{Evaluation protocols:} Since ground-truth videos are not available for use in evaluating generated videos for the motion transfer task, we follow the FOMM~\cite{siarohin2020first} evaluation protocol and take self-reconstruction as a proxy task to quantitatively evaluate the proposed method. More specifically, an input video is reconstructed from the appearance representation of its first frame and the motion flow of the entire videos according to Eqn.~(\ref{global flow}). The same four different metrics as in~\cite{siarohin2020first} are used for evaluation.
\begin{itemize}[itemsep=0pt]
    \item $\mathcal{L}_1$. The average $\mathcal{L}_1$ distance between the pixel values of generated and ground-truth video frames.
    \item Average Keypoint Distance (AKD). This metric computes the average keypoint distance between generated and ground-truth video frames. It is designed to evaluate the pose quality of the generated video frames.
    \item Missing Keypoint Rate (MKR). For human body datasets, we further report MKR, which represents the percentage of keypoints that are not detected in generated video frames but are localized in the ground truth video frames. 
    \item Average Euclidean Distance (AED). This metric is designed to assess the identity quality of generated video frames based on specific feature representations; 
    in the feature space, the average Euclidean distance between generated and ground-truth video frames is computed.   
\end{itemize}

\noindent\textbf{Implementation details:} Seen in the supplementary.

\begin{figure*}[t]
\begin{center}
  \includegraphics[width=1\linewidth]{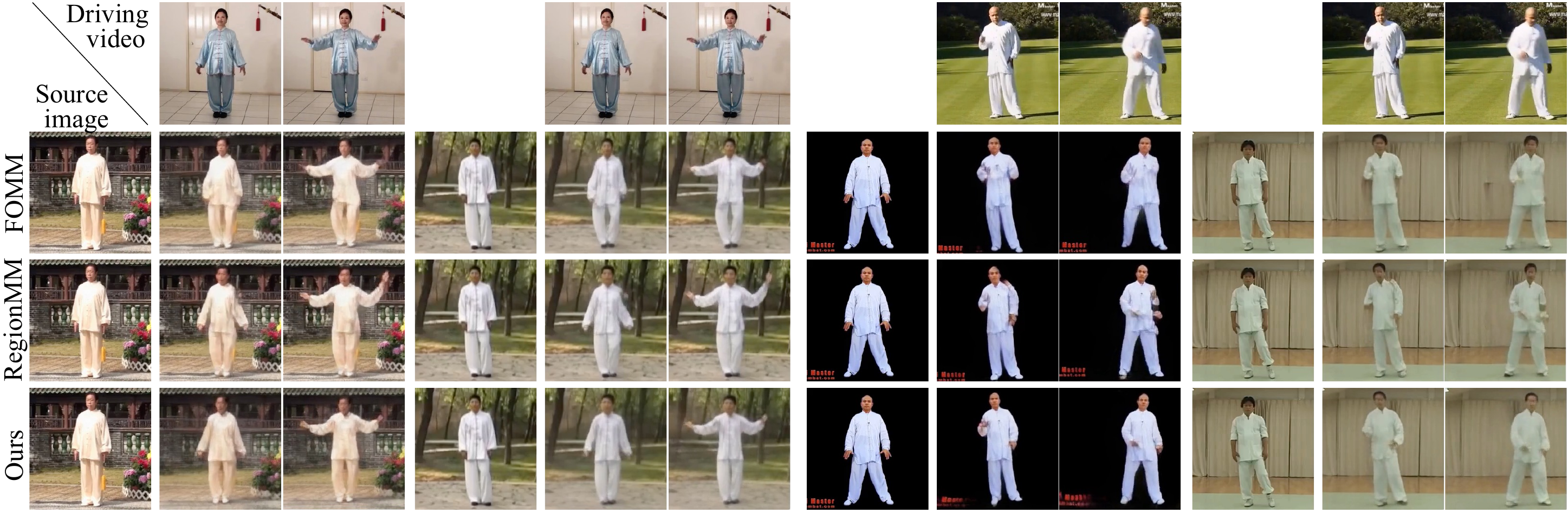}
  \caption{Qualitative comparisons on cross-identity motion transfer. We present four source identities driven by two videos from the TaiChiHD dataset. It can be seen that our method generally synthesizes the most structure-stable results.}
  \label{fig:animation compare}
\end{center}
\vspace{-0.5cm}
\end{figure*}

\begin{table*}[]
\begin{center}
\resizebox{\textwidth}{!}
{
\begin{tabular}{c|ccc|ccc|ccc|ccc|c}
\hline
\multicolumn{1}{l|}{} & \multicolumn{3}{c|}{TaiChiHD $256\times256$}                   & \multicolumn{3}{c|}{TaiChiHD $512\times512$}                   & \multicolumn{3}{c|}{Fashion}                                     & \multicolumn{3}{c|}{VoxCeleb1}             & MGIF \\
\multicolumn{1}{l|}{} & L1             &  (AKD, MKR)               & AED              & L1             &  (AKD, MKR)               & AED              & L1    &  (AKD, MKR)      & AED                                     & L1        & AKD         & AED              & L1   \\ \hline
Monkey-Net            & 0.077          &  (10.798, 0.059)          & 0.228            & -          & -          & -            & - & -   & -                                   & 0.049     & 1.89       & 0.199            & -    \\
FOMM                  & 0.057          &  (6.649, 0.036)           & 0.172            & 0.065          &  (15.08, 0.061)          & 0.202            & 0.013 &  (1.142,0.005)   & 0.059                                   & 0.041     & 1.28       & 0.133            & 0.0224\\
RegionMM              & 0.048          &  (5.246, 0.024)           & 0.150            & 0.057          &  (11.97, 0.028)          & 0.166            & \textbf{0.011} &  (1.187,0.005) & 0.056                                   & 0.040     & 1.28       & 0.133            & 0.0206\\ \hline
Ours DAM              & 0.045          &  (5.102, 0.024)          & 0.150            & 0.054           &  (10.83, 0.032)          & 0.158            & \textbf{0.011} &  (1.116, 0.005) & 0.055                                   & 0.040     & 1.26       & 0.130            & 0.0207 \\
Ours HDAM             & \textbf{0.044} &  (\textbf{4.790, 0.021})  & \textbf{0.146}    & \textbf{0.053} &  (\textbf{10.19, 0.027}) & \textbf{0.156}   & \textbf{0.011} &  (\textbf{1.041, 0.004}) & \textbf{0.054}        & \textbf{0.039} & \textbf{1.24} & \textbf{0.124} &\textbf{0.0201} \\  \hline
\end{tabular}
}
\caption{Quantitative comparisons on the self-reconstruction task. We present results on four benchmarks; here, a lower score is preferred for all metrics. For fair comparison, motion anchors are set to 10 for all methods.}
\label{tab1}
\end{center}
\vspace{-0.7cm}
\end{table*}

\begin{table}[]
\begin{center}
\begin{tabular}{c|ccc}
\hline
\multicolumn{1}{l|}{}            & TaiChiHD    & Fashion     & Voxceleb1   \\ \hline
\multicolumn{1}{c|}{Ours vs FOMM}        & 91.6\% & 76.0\% & 54.2\% \\
\multicolumn{1}{c|}{Ours vs RegionMM}    & 59.1\% & 66.6\% & 66.0\% \\ \hline
\end{tabular}
\caption{User preferences favoring our approach.}
\label{tab2}
\end{center}
\vspace{-0.65cm}
\end{table}

\subsection{Comparison with Existing Methods}

We compare our method with two recent model-free motion transfer methods: FOMM~\cite{siarohin2020first} and RegionMM\cite{PCAMotion}. 

\noindent\textbf{Quantitative results:} The comparisons are summarized in Table~\ref{tab1}. We can observe that our proposed HDAM approach generally achieves the best performance on all evaluation metrics. In particular, the fact that our $\mathcal{L}_1$ score is the lowest reflects the good quality of the videos generated by our method. Moreover, the improvement to AKD and MKR indicates that our method achieves good motion transfer, while the improved AED also reflects the appearance quality of the videos generated using our method. 

In more detail, compared to the FOMM method, we achieve a notable improvement on the TaiChiHD, FashionVideo and MGIF dataset, while also gaining better results on the VoxCeleb dataset. This clearly proves the effectiveness of using deformable anchor models to regularize motion anchors. Moreover, the fact that our work outperforms the most recent related work, RegionMM, further proves the advantages of modeling object structure; notably, this superiority also holds in the case of higher-resolution inputs. We further note that the improvements on the VoxCeleb dataset are not as significant as those on the TaiChiHD and MGIF datasets. This is possibly because the structures of the human face are relatively simple, while the human body consists of multiple joints and movable parts, meaning that its motion are usually quite complicated. These results reflect that the deformable anchor model helps to transfer motion on various objects, especially those with complicated structures, which also validates the motivation of this work. 

\noindent\textbf{User study:} We conduct a user study for cross-identity motion transfer. More specifically, we prepare 50 concatenated results consisting of a source frame, driving videos and videos generated by FOMM, RegionMM and our method; the synthesized videos are placed in random order in each of the concatenated videos. Fifty participants are asked to rank the three videos based on the appearance preservation and transferred motion. As Table~\ref{tab2} shows, participants clearly identified our videos as being of higher quality than the synthesized videos produced by existing methods.

\noindent\textbf{Qualitative results:} We additionally present examples of the videos generated by the three methods in Fig.~\ref{fig:animation compare}. Generally speaking, FOMM often synthesizes an abnormal body shape or an incorrect motion from the driving video. Moreover, while RegionMM is able to roughly depict the motion contained in the driving video, it may also fail to capture more detailed structural information, leading to obvious artifacts (\eg, the lost or weirdly warped human arms). By contrast, our method is generally able to capture the motion details well and produces more stable results. More qualitative results are provided in the supplementary material.

\subsection{Ablation Study} \label{4.3}

We next conduct an ablation study to analyze the impact of our proposed components. Specifically, we study two variants of our proposed approach: 1) the basic deformable anchor model in Section~\ref{3.2} (referred to as ``Ours (DAM)"), and 2) the hierarchical deformable anchor model in Section~\ref{3.3} (referred to as ``Ours (HDAM)"). We further employ FOMM in which no deformable anchor model is used, as a baseline for comparison.

As seen in Table~\ref{tab1}, we conduct experiments on the TaiChiHD dataset and analyze the results. We observe that \textit{Ours (DAM)} achieves considerable improvements relative to the baseline FOMM,  confirming the validity of exploiting object structures with a deformable anchor model in order to improve motion transfer. Moreover, by introducing the hierarchical deformable anchor model, \textit{Ours (HDAM)}) achieves further improvements.

\begin{figure}[t]
\begin{center}
   \includegraphics[width=0.88\linewidth]{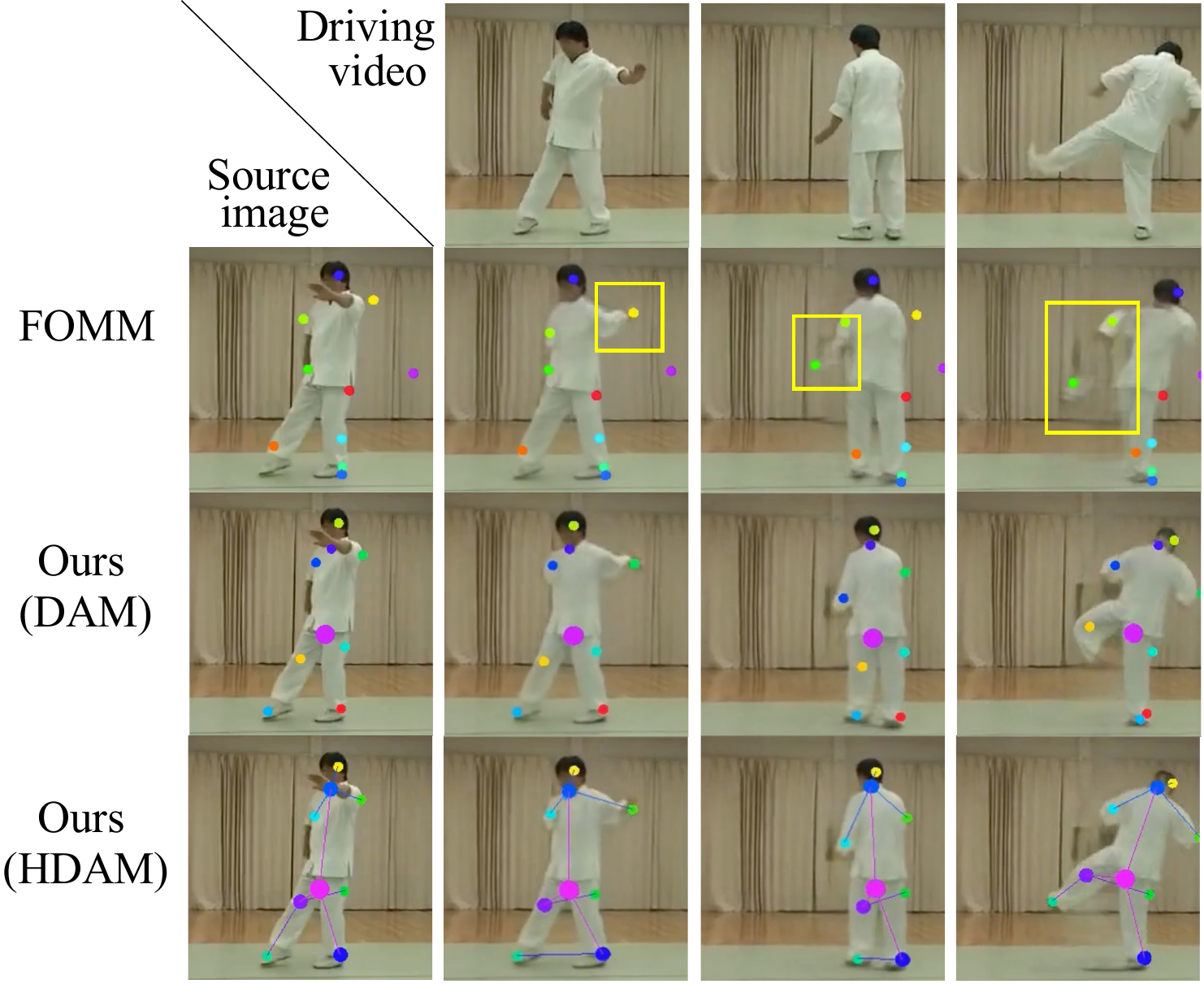}
   \caption{Qualitative ablation study. We visualize the latent root anchor as the largest dot and denote the intermediate anchors using medium-sized dots; correspondingly, the smallest dots represent motion anchors. Adjacent anchors are connected through straight lines according to the max attention weight, which reflects the constraint relation. Note that all learned intermediate anchors are overlapped with a motion anchor in this dataset; further analysis of this is provided in the supplementary material.}
   \label{fig:ablation}
 \vspace{-0.54cm}
\end{center}
\end{figure}

In Fig.~\ref{fig:ablation}, we present qualitative examples of our ablation study to reveal how our method works. We draw predicted anchors on generated frames to facilitate detailed analysis. As can be seen from the figure, FOMM generally fails to capture the local structure of the human body (such as hands and legs) due to the incorrectly aligned motion anchors; by contrast, \textit{Ours(DAM)} can synthesize a relatively complete object structure, reflecting the effectiveness of DAM in constraining the object structure. Furthermore, \textit{Ours(HDAM)} generally learns the meaningful structure and synthesizes high-quality results while capturing stable and complete structure information, which further verifies the superiority of modeling the hierarchical object structure. 

\subsection{Parameter Analysis} \label{4.4}

To validate the robustness of the proposed method, we study the influence of different hyper-parameter settings in this section. Specifically, we examine two important hyper-parameters in our model, namely the number of motion anchors and the number of intermediate anchors. We conduct experiments on the TaiChiHD and MGIF datasets to perform this analysis. The quantitative results are obtained from the \textit{Ours (HDAM)} model. As seen in Table~\ref{tab3}, our method continues to improve as more motion anchors are involved, with a large improvement visible when moving from 5 to 10 motion anchors and a relatively small one from 10 to 20 motion anchors. Moreover, our method generally works well with $2\sim6$ intermediate anchors, as can be seen from Table~\ref{tab4}. Our observations further suggest that, when more intermediate anchors are involved, the HDAM model tends to learn that some of them are meaningless or overlapping with other intermediate anchors; we explain that the object structure is overfitted in these situations. Overall, our method generalizes well to these hyper-parameters.

\subsection{Structure Visualizations} \label{4.5}

To understand the proposed methods in more depth, we visualize the predicted hierarchical anchors of video frames on different datasets in Fig.~\ref{fig:introduction} and Fig.~\ref{fig:ablation}; more qualitative results are provided in the the supplementary material. As is evident from the results, the learned root anchor is always located at the object centroid regardless of its identity or background; moreover, intermediate root anchors are often located at different local regions of an object, which enables them to capture more detailed motions of the parts in question. Note that in our hierarchical model, as seen in the fourth row of Fig.~\ref{fig:ablation}, when different motions occur, motion anchors can be regularized by different intermediate anchors according to the attention weights in Eqn.~(\ref{second order loss}). This reflects the ability of our HDAM model to flexibly constrain the motion structures according to the varying motions in the dataset. In summary, this star-like motion structure learned by our deformable anchor model exhibits a strong ability to model stable motions between images.

\begin{table}[]
\begin{center}
\begin{tabular}{c|c|ccc}
\hline
\multicolumn{1}{l|}{}  & MGIF            & \multicolumn{3}{c}{TaiChiHD}                       \\
\multicolumn{1}{l|}{}  & L1              & L1             & (AKD, MKR)               & AED                     \\ \hline
5                      & 0.0235          & 0.048          & (5.730, 0.028)          & 0.159                   \\
10                     & 0.0201          & 0.044          & (4.790, 0.021)          & 0.146  \\
20                     & \textbf{0.0185}          & \textbf{0.043}          & \textbf{(4.615, 0.018)}          & \textbf{0.138}                   \\ \hline
\end{tabular}
\caption{Quantitative performance with different number of motion anchors. We assess performance at 5, 10, 20 motion anchors respectively. The number of intermediate anchors is fixed at 3.}
\label{tab3}
\end{center}
\vspace{-0.5cm}
\end{table}

\begin{table}[]
\begin{center}
\begin{tabular}{c|c|ccc}
\hline
\multicolumn{1}{l|}{}                          & MGIF            & \multicolumn{3}{c}{TaiChiHD}                       \\
\multicolumn{1}{l|}{}                          & L1              & L1              & (AKD, MKR)               & AED                     \\ \hline
2                                              & 0.0202          & 0.045           &  (\textbf{4.763, 0.021})          & \textbf{0.146}                   \\
3                                              & 0.0201          & \textbf{0.044}          & (4.790, \textbf{0.021})          &\textbf{0.146}  \\
4                                              & 0.0201          & \textbf{0.044}          & (4.836, 0.022)          & \textbf{0.146}                   \\
5                                              & \textbf{0.0199}          & 0.045          & (4.926, 0.023)          & \textbf{0.146}                   \\
6                                              & 0.0200          & \textbf{0.044}          & (4.792, 0.022)          & \textbf{0.146}                   \\ \hline
\end{tabular}
\caption{Quantitative performance with different numbers of intermediate anchors. We tune $2\sim6$ intermediate anchors respectively. The number of motion anchors is fixed at 10.}
\label{tab4}
\end{center}
\vspace{-0.6cm}
\end{table}


\section{Conclusion}
This paper proposes a novel structure-aware motion transfer approach with deformable anchor model. In DAM, the latent root anchor is designed to constrain the motion anchors. We then explore the intermediate latent root anchors to build hierarchical DAM, leading to structure-stable motion transfer and yielding the best performance (both qualitatively and quantitatively) relative to existing benchmarks. 
We further interpret our method through insightful an ablation study and validate the robustness of our method to different hyper-parameter settings. 

\noindent\textbf{Societal impact and limitations:} Motion transfer techniques could be misused for generating fake videos, which might bring negative societal impact. People should be cautious and get authorized when manipulating videos using these techniques. Moreover, while we demonstrate state-of-the-art performance, the results are not perfect. Some artifacts can still be observed when there exists occlusion, large motion, complex background, \etc. We will study these issues in the future. 

\noindent\textbf{Acknowledgement:} This work is partially supported by Alibaba Group through Alibaba Innovation Research Program.

{\small
\bibliographystyle{ieee_fullname}
\bibliography{egbib}
}

\end{document}